\documentclass[final]{article} 
\newif\ifpreprint
\preprinttrue

\usepackage[nonatbib]{neurips_2020}

\usepackage[utf8]{inputenc} 
\usepackage[T1]{fontenc}    
\usepackage{hyperref}       
\usepackage{url}            
\usepackage{booktabs}       
\usepackage{amsfonts}       
\usepackage{nicefrac}       
\usepackage{microtype}      
\usepackage{tikz}
\usepackage{times}
\usepackage{epsfig}
\usepackage{graphicx}
\usepackage{amsmath}
\usepackage{amssymb}
\usepackage{multirow}
\usepackage{float}
\usepackage{subcaption}
\usepackage{stmaryrd}
\usepackage{xspace}
\usepackage{wrapfig}

\usepackage[style=numeric-comp,date=year,doi=false,isbn=false,url=false,eprint=false,maxbibnames=8]{biblatex}
\graphicspath{ {images/} }

\usepackage{ifthen}
\usepackage{xcolor}
\definecolor{todo}{rgb}{1,0,0}
\definecolor{todoo}{rgb}{1,0,0}
\newcommand{\todo}[1]{
\color{todo}
\ifthenelse { \equal {#1} {} }  
{ \textbf{todo} }   
{ \textbf{todo: #1} }   
\color{black}
}

\newcommand{\transform}{\ensuremath{\Phi}}

\newcommand{\domain}{\ensuremath{\Omega}}

\newcommand{\dist}{\ensuremath{D}}
\newcommand{\reg}{\ensuremath{R}}
\newcommand{\loss}{\ensuremath{L}}

\DeclareMathOperator{\mse}{MSE}
\DeclareMathOperator{\ncc}{NCC}
\DeclareMathOperator{\deepsim}{DeepSim}

\DeclareMathOperator{\vgg}{VGG}
\newcommand{\I}{\ensuremath{\mathbf{I}}}
\newcommand{\J}{\ensuremath{\mathbf{J}}}

\newcommand{\Ss}{\ensuremath{\mathbf{S}}}

\newcommand{\pcoord}{\ensuremath{\mathbf{p}}}

\author{%
	Steffen Czolbe ~~~~~~~~~~~~~~~~~~~~ Oswin Krause\\
	Department of Computer Science\\
	University of Copenhagen\\
	\texttt{\{per.sc, oswin.krause\}@di.ku.dk} \\
	\And
	\And
	Aasa Feragen \\
	DTU Compute\\
	Technical University of Denmark\\
	\texttt{afhar@dtu.dk} \\
}

\title{DeepSim: Semantic similarity metrics for learned image registration}

\begin{document}
	\maketitle
	
	\begin{abstract}
		We propose a semantic similarity metric for image registration. Existing metrics like euclidean distance or normalized cross-correlation focus on aligning intensity values, giving difficulties with low intensity contrast or noise. Our semantic approach learns dataset-specific features that drive the optimization of a learning-based registration model. Comparing to existing unsupervised and supervised methods across multiple image modalities and applications, we achieve consistently high registration accuracy and faster convergence than state of the art, and the learned invariance to noise gives smoother transformations on low-quality images. 
	\end{abstract}
	
	\section{Introduction}
	Deformable registration is a fundamental preprocessing tool in medical imaging, where the goal is to find anatomical correspondences between images and derive geometric transformations $\transform$ to align them. Most algorithmic and deep learning-based methods optimize alignment via a similarity measure $\dist$ and a $\lambda$-weighted regularizer $\reg$, combined in a loss function
	\begin{equation} \label{eq:reg-loss}
	\loss(\I, \J, \transform) = \dist(\I \circ \transform, \J) + \lambda \reg(\transform) \enspace .
	\end{equation}
	The similarity metric $D$ assesses the alignment and strongly influences the result.	Pixel-based similarity metrics like euclidean distance ($\mse$) and patch-wise normalized cross-correlation ($\ncc$) are commonly used in both algorithmic~\cite{Avants2011, Avants2008, faisalBeg2005, Rueckert1999, thirion1998, Vercauteren2007} and deep learning based~\cite{Alven2019, balakrishnan2019voxel, dalca2018unsuperiveddiff, Dalca2019, DeVos2019, Liu2019, Lee2019a, Hu2019, Hu2019a, Yang2017, Xu2019} image registration. Typically, the similarity measure for a task is selected as the best out of a small set of metrics, with no guarantee that one of the metrics is suitable for the data.
	
	The shortcomings of pixel-based similarity metrics have been studied substantially in the image generation community \cite{Zhang2018}, where the introduction of deep similarity metrics approximating human visual perception has improved the generation of photo-realistic images \cite{Czolbe2020, hou2017deep}. As registration models are generative models \cite{dalca2018unsuperiveddiff}, we expect these similarity metrics to improve registration as well. Current attempts at using learned similarity metrics for image registration require ground truth transformations \cite{Haskins2019a} or limit the input to the registration model \cite{Lee2019a}. 
	
	We propose a data-driven similarity metric for image registration based on the alignment of semantic features. We learn semantic filters of our metric on the dataset, use it to train a registration model, and validate our approach on three biomedical datasets of different image modalities and applications. Across all datasets, our method achieves consistently high registration accuracy, outperforming even metrics utilizing supervised information. Our models converge faster and learn to ignore noisy image patches, leading to smoother transformations on low-quality data.

	\begin{figure}
		\centering
		\def\svgwidth{0.9 \linewidth}
		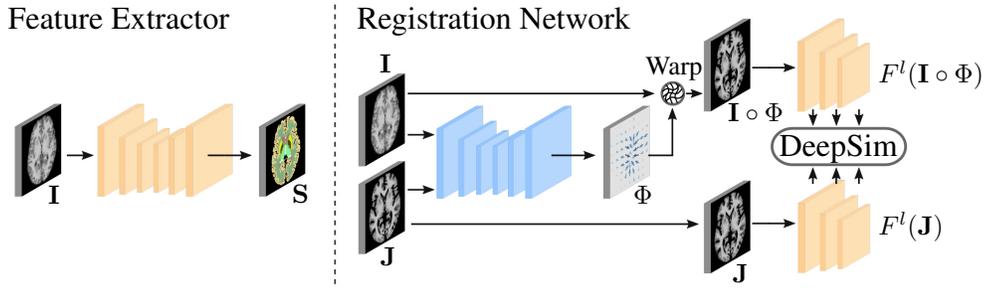
		\caption{Two-step training: First, the Feature Extractor (yellow) is trained on a segmentation task. Next, its weights are frozen and used in the loss computation of the registration network (blue).}
		\label{fig:model}
	\end{figure}
	
	\section{A deep similarity metric for image registration}
	To align areas of similar semantic value, we propose a similarity metric based on the agreement of semantic feature representations of two images. Semantic feature maps are obtained by a \textit{feature extractor} to be tuned on a surrogate segmentation task. To capture alignment of both localized, concrete features, and global, abstract ones, we calculate the similarity at multiple layers of abstraction.
	
	Concretely, given a set of feature-extracting functions $F^l \colon \mathbb{R}^{\domain \times C} \to \mathbb{R}^{\domain_l \times C_l}$ for $L$ layers, we define
	\begin{align}
	\deepsim(\I \circ \transform, \J) = \frac{1}{L} \sum_{l=1}^{L} \frac{1}{| \domain_l |} \sum_{{\pcoord} \in \domain_l}
	\frac{\big< F^l_{\pcoord}(\I \circ \transform), F^l_{\pcoord}(\J)\big>}{\Vert  F^l_{\pcoord}(\I \circ \transform) \Vert \Vert  F^l_{\pcoord}(\J) \Vert} \enspace ,
	\end{align}
	where $F^l_{\pcoord}(\J)$ denotes the $l^{th}$ layer feature extractor applied to image $\J$, at spatial coordinate $\pcoord$. It is a vector of $C_l$ output channels, and the spatial size of the $l^{th}$ feature map is denoted by $| \domain_l |$. The metric is influenced by the neighbourhood of a pixel, as  $F^l$ composes convolutional filters with increasing receptive area of the composition. Note that the formulation via cosine similarity is similar to the classic $\ncc$ metric, which can be interpreted as the squared cosine-similarity between two zero-mean vectors of patch descriptions.
	
	\paragraph{Feature extraction}
	To aid registration, the functions $F^l(\cdot)$ should extract features of semantic relevance for the registration task, while ignoring noise and artifacts. We achieve this by training the feature extractor on a supplementary segmentation task, as segmentation models excel at learning relevant kernels for the data while attaining invariance towards non-predictive features like noise. The obtained convolutional filters act as feature extractors for $\deepsim$, see also Figure~\ref{fig:model}.
	

	\section{Experiments}
	
	\begin{wrapfigure}{r}{0.65\textwidth}
		\vspace{-10ex}
		\includegraphics[width=1\linewidth]{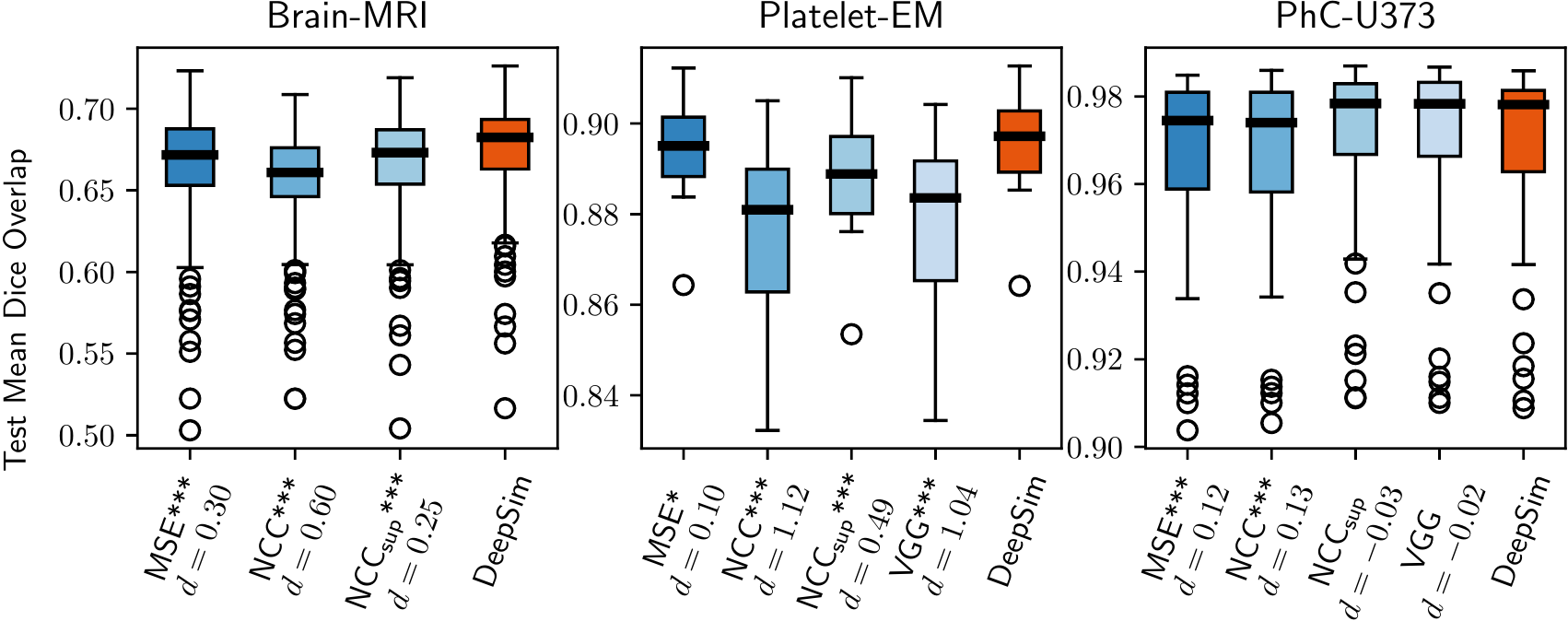}
		\caption{Quantitative comparison of similarity metrics. Stars indicate p-test significance level. Effect size given by Cohen's d.}
		\label{fig:testscore}
		\vspace{-1ex}
	\end{wrapfigure}
	We compare registration models trained with $\deepsim$ to the baselines $\mse$, $\ncc$, $\ncc_{\text{sup}}$ ($\ncc$ with supervised information \cite{balakrishnan2019voxel}), and $\vgg$ (a VGG-net based metric common in image generation and similar to our method \cite{hou2017deep}). Figure~\ref{fig:model} shows our model architecture. For both registration and segmentation we use U-nets \cite{Ronneberger2015}. The registration network predicts the transformation $\transform$ based on two images $\I, \J$. A spatial transformer module \cite{jaderberg2015spatial} applies $\transform$ to obtain the morphed image $\I \circ \transform$. The loss is given by Eq. \ref{eq:reg-loss}; we choose the diffusion regularizer for $R$ and tune hyperparamter $\lambda$ on the validation sets.
	
	To show that our approach is applicable to a large variety of registration tasks, we validate it on three 2D and 3D datasets of different image modalities: T1-weighted \textit{Brain-MRI} scans \cite{di2014autism, lamontagne2019oasis}, human blood cells of the \textit{Platelet-EM} dataset  \cite{Quay2018}, and cell-tracking of the \textit{PhC-U373} dataset \cite{Ulman2014, Ulman2017}. Each dataset is split into a train, validation, and test section.

	\begin{figure}
		
		\begin{subfigure}[b]{.78\textwidth}
			\centering
			\begin{tikzpicture}
			\node[anchor=south west,inner sep=0] (image) at (0,0) {%
				\includegraphics[width=1\textwidth]{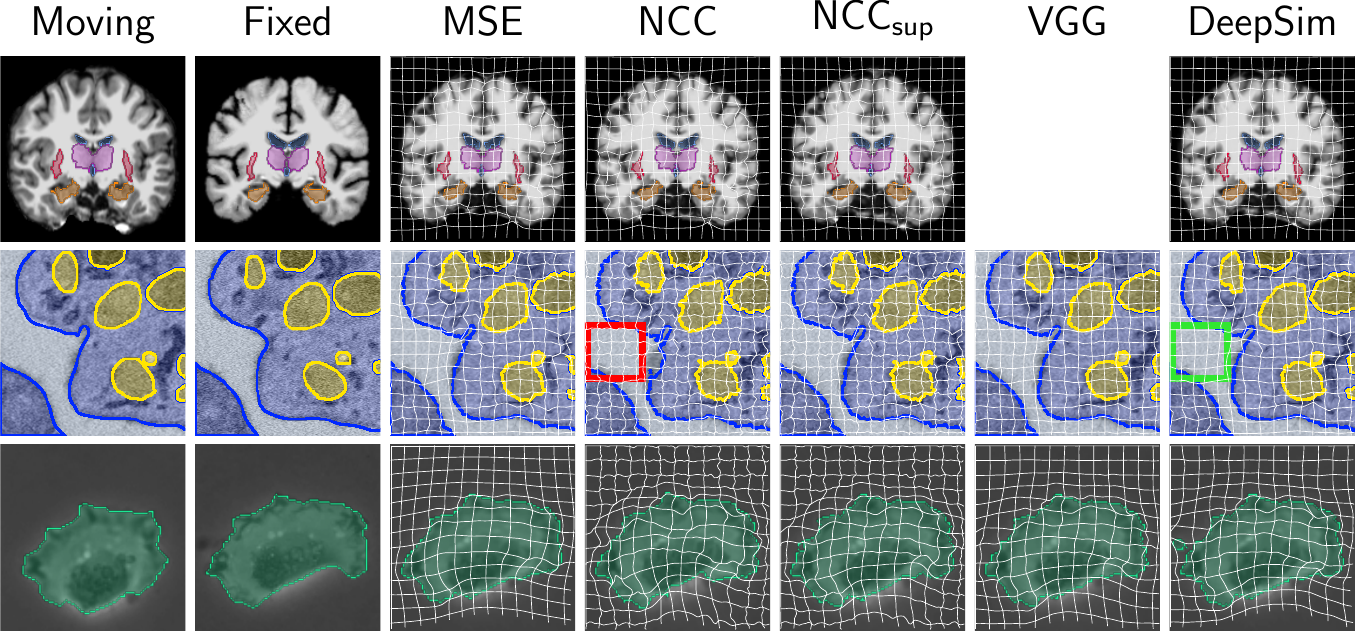}
			};
			\draw (7.84,4.62) -- (9.31,4.62) -- (9.31,3.15) -- (7.84,3.15) -- (7.84,4.62);
			\node[align=center, scale=0.75] at (8.575,3.885) { N/A for \\ 3D Data};
			\end{tikzpicture}
			\caption{}
			\label{fig:samples}
		\end{subfigure}%
		\hfill
		\begin{subfigure}[b]{.18\textwidth}
			\centering
			\includegraphics[width=0.805\linewidth]{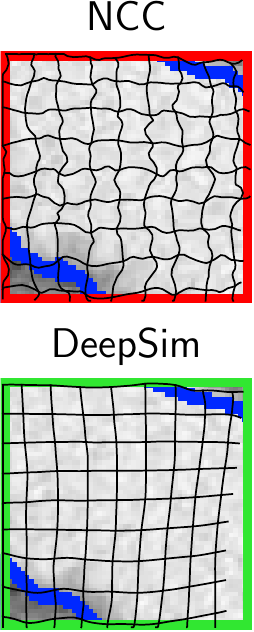}
			\caption{}
			\label{fig:grids}
		\end{subfigure}
		\caption{(a) Qualitative comparison, (b)~Detail view of highlighted areas. Select segmentation classes annotated color. The transformation is visualized by morphed grid-lines. }
	\end{figure}
	
	\paragraph{Registration accuracy \& convergence}
	We measure the mean S{\o}rensen Dice coefficient on the unseen test-set in Figure~\ref{fig:testscore}, and test for statistical significance of the result with the Wilcoxon signed rank test for paired samples. Our null hypothesis for each similarity metric is that the model trained with $\deepsim$ performs better. We test for a statistical significance levels of ${p^{*} = 0.05, p^{**} = 0.01, p^{***} = 0.001}$. We further measure the effect size with Cohen's d, and label the metrics accordingly in Figure~\ref{fig:testscore}.
	Models trained with our proposed $\deepsim$ rank as the best on the Brain-MRI and Platelet-EM datasets, with strong statistical significance. On the \mbox{PhC-U373} dataset, all models achieve high dice-overlaps of $>0.97$. $\deepsim$ converges faster than the baselines, especially in the first epochs of training.

	\vspace{-1ex}
	\paragraph{Qualitative examples \& transformation grids}
	We plot the fixed and moving images $\I, \J$ and the morphed image $\I \circ \transform$ for each similarity metric model in Figure~\ref{fig:samples}, and a more detailed view of a noisy patch of the Platelet-EM dataset in Figure~\ref{fig:grids}. The transformation is visualized by grid-lines, which have been transformed from a uniformly spaced grid. On models trained with the baselines, we find strongly distorted transformation fields in noisy areas of the images. In particular, models trained with $\ncc$ and $\ncc_{\text{sup}}$ produce very irregular transformations, despite careful tuning of the regularization-hyper-parameter. The model trained by $\deepsim$ is more invariant towards the noise.

	\section{Discussion \& Conclusion}
	Registration models trained with $\deepsim$ achieve high registration accuracy across multiple datasets, leading to improved downstream analysis and diagnosis in medical applications. 
	The consistency of our proposed metric makes testing multiple traditional metrics unnecessary; instead of empirically determining whether $\mse$ or $\ncc$ captures the characteristics of a data-set best, we can use $\deepsim$ to learn the relevant features from the data. 
	
	The analysis of noisy patches in Figure~\ref{fig:grids} highlights a learned invariance to noise. The pixel-based similarity metrics are distracted by artifacts, leading to overly-detailed transformation fields. $\deepsim$ does not show this problem. While smoother transformation fields can be obtained for all metrics by strengthening the regularizer, this would negatively impact the registration accuracy of anatomically significant regions. Accurate registration of noisy, low-quality images allows for shorter acquisition time and reduced radiation dose in medical applications.
	
	
	$\deepsim$ is a general metric, applicable to image registration tasks of all modalities and anatomies. Beyond the presented datasets, the good results on low-quality data let us hope that $\deepsim$ will improve registration accuracy in the domains of lung CT and ultrasound, where details are hard to identify, and image quality is often poor. We further emphasize that the application of $\deepsim$ is not limited to deep learning. Algorithmic image registration uses a similar optimization framework, where a similarity-based loss is minimized via gradient descent-based methods. $\deepsim$ can be applied to drive algorithmic methods, improving their performance by aligning deep, semantic feature embeddings.

	\section*{Broader impact}
	The broader impact of our work is defined by the numerous applications of medical image registration. Common applications are in neuroscience \cite{balakrishnan2019voxel}, CT-imaging of lungs and abdomen \cite{DeVos2019}, as well as for fusion and combination of multiple modalities \cite{Haskins2019a}. 
	
	The deep learning approach to image registration utilized in this work can achieve impressive results across a wide variety of tasks, but this often comes at the cost of training models on specialized hardware for extensive periods. This energy-intensive workload may raise carbon emissions, the primary contributor to climate change \cite{Anthony2020}. We hope that by presenting a method for learning a semantic similarity metric from the data, we make excessive testing of other loss functions unnecessary. This can reduce the amount of model configurations to be tested in the development of deep learning methods, contributing to a lower environmental impact of the image registration community.

	\printbibliography
	
\end{document}